\def\year{2020}\relax
\documentclass[letterpaper]{article} 

\usepackage{aaai20}  
\usepackage{times}  
\usepackage{helvet} 
\usepackage{courier}  
\usepackage[hyphens]{url}  
\usepackage{graphicx} 
\usepackage{graphicx}  
\usepackage{microtype}

\usepackage{latexsym}
\usepackage{multirow}
\usepackage{hyperref}
\usepackage{booktabs}
\usepackage{graphicx}
\usepackage{amsmath}
\usepackage{tablefootnote}
\usepackage{xspace}
\usepackage{dsfont}
\usepackage{balance}
\usepackage{url}
\usepackage[T1]{fontenc}
\usepackage[utf8]{inputenc}
\usepackage{fixltx2e}
\usepackage{amsmath,amssymb}

\usepackage{comment}
\usepackage{amsmath}

\newcommand{\sentref}{x}
\newcommand{\senthyp}{\hat{x}}
\newcommand{\context}{c}
\newcommand{\generation}{G}
\newcommand{\Reference}{R}

\newcommand{\metric}[1]{\textsc{#1}\xspace}

\newcommand{\bleu}{\metric{Bleu}}

\newcommand{\meteor}{\metric{Meteor}}
\newcommand{\rouge}{\metric{Rouge}}

\newcommand{\ruse}{\metric{Ruse}}

\newcommand{\perplexity}{\metric{perplexity}}
\newcommand{\bleurt}{\metric{bleurt}}

\newcommand{\rocstory}{ROCStories}
\newcommand{\imdbCon}{Large Movie Review Conditional}
\newcommand{\imdbUn}{Large Movie Review Unconditional}
\newcommand{\coco}{COCO Image Captions}

\newcommand{\roberta}{RoBERTa}
\newcommand{\bert}{BERT}

\newcommand{\Var}{\mathrm{Var}}

\newcommand{\moverscore}{\metric{MoverScore}}
\newcommand{\bertscore}{\metric{BERTScore}}

\newcommand{\method}{\metric{Perception Score}}

\DeclareMathOperator*{\argmax}{arg\,max}

\newcommand{\citet}[1]{\citeauthor{#1} \shortcite{#1}}
\newcommand{\citep}{\cite}

\title{Perception Score, A Learned Metric for Open-ended Text Generation Evaluation}
\author{
Jing Gu\textsuperscript \,\,\, Qingyang Wu \,\,\, Zhou Yu \\
University of California, Davis \\
\{jkgu,wilwu,joyu\}@ucdavis.edu 
1 Shields Ave, Davis, California 95616\\ 
}

\begin{document}

\maketitle

\begin{abstract}
    Automatic evaluation for open-ended natural language generation tasks remains a challenge. Existing metrics such as \bleu show a low correlation with human judgment. We propose a novel and powerful learning-based evaluation metric: \method.
The method measures the overall quality of the generation and scores holistically instead of only focusing on one evaluation criteria, such as word overlapping. Moreover, it also shows the amount of uncertainty about its evaluation result. By connecting the uncertainty, \method gives a more accurate evaluation for the generation system. \method provides state-of-the-art results on two conditional generation tasks and two unconditional generation tasks.  
\end{abstract}

\section{Introduction}
\label{sec:intro}

With the recent advances in natural language generation (NLG) tasks, automatic evaluation has drawn more attention from the research community. However, the evaluation of an advanced generation model faces obstacles with the shortcomings of existing metrics, especially in open-ended text generation tasks.

Existing metrics such as \bleu and \bleurt have been proved useful in various text generations evaluation tasks. Usually, these metrics measure the similarity between the reference and the candidate in different standards.
However, when it comes to open-ended tasks, these metrics will become less effective.

N-gram overlapping based metrics such as \bleu \citep{bleu_metric} and \rouge \citep{rouge_metric} could be the most widely used ones across different NLG tasks. These metrics only consider word-form variation and often fail to capture the real semantic meaning which the reference or the candidate conveys. A candidate sentence that shares similar semantic meaning with the reference will be scored unfairly low under n-gram matching. As a result, these metrics inevitably show poor correlation with human judgment \citep{how-not-evalute-dialog,why-new-nlg-metric,price-of-debias-metric}.
Recently, different neural network-based metrics are also proposed towards text generation evaluation. \bertscore \citep{bertscore} replace the hard n-gram matching with a soft similarity of context embedding. \moverscore \citep{moverscore} word embeddings of a pre-trained model to find the semantic similarity via Word Mover's Distance. Both methods utilize prior knowledge in a large pre-trained neural model. 
However, the semantic meaning of generation and reference could be context-dependent, especially in open-ended generation task. Furthermore, for those metrics which claim to capture the semantic meaning, can they really measure the quality of a sentence? For example, how do they measure a story that is enchanting but also reasonable, or how do they understand a poem is striking but also rhyming?
It requires more than the surface of words to recognize and comprehend in open-ended generation tasks.
To truly fathom the quality of generations, we need to transcend the limits of using the superficial understanding of a language.

In the creed of building a universal and straightforward evaluation metric for open-ended text generation tasks, we propose \method, a system-level automated evaluation metric that diffuses evaluation onto the multi-dimensional space and assigns a single holistic score based on its overall quality to system-generated text with the reference as goal standard.
This gives us the most direct observation of the quality through the comparisons of the reference and the generated text. By using a large pre-training model, \method learns to measures the difference between the generations and the distribution of references.
Moreover, \method also shows data uncertainty estimation and model uncertainty estimation when scoring generated text. \method provides a more accurate overall evaluation for a generation system by focusing more on the generation that is scored with confidence.
We validate \method with extensive experiments on four tasks, including two conditional generation tasks, i.e., \rocstory{} and \imdbCon{}, and two unconditional tasks, i.e., \imdbUn{} and \coco{}.
\method shows stronger co-relation with human judgement than other metrics including \bleu, \perplexity, \bertscore and \bleurt. Besides, it show more robustness than other metrics under several adversarial attacks.
\section{Related Works}
\label{sec:related_works}

Evaluation is an essential topic in natural language generation. While human evaluation is probably the most accurate metric for generation evaluation \citep{evaluation-survey}, it is usually time-consuming and expensive. Researchers have proposed different types of automated evaluation metrics to replace human evaluation.

$n$-gram matching is the most used evaluation method in the natural language generation task. In the machine translation task, \bleu is a common metric to evaluate the similarity between candidates and references in word surface level. \meteor is also an automatic metric for machine translation evaluation which considers surface forms, stemmed forms and meanings. \rouge \citep{rouge_metric} is commonly used in summarization evaluation task. 
The performance of $n$-gram matching metric only considers similarity in word surface.

\perplexity is another commonly used metric in open-ended generation tasks such as chit-chat tasks. It has been proven to be a proper measurement of the quality of a language model but can not directly reflect the generation sentence quality.

Recently, context embedding in the large pre-trained model is also used for evaluation. \bertscore utilizes semantic embeddings in large pre-trained neural models for each token, then applies cosine similarity and greedy matching to maximize the similarity score between references and candidates. \moverscore combines the contextual embeddings from a pre-trained model and Word Mover’s Distance to evaluate text generation. These metric utilizes large pre-trained models to could evaluate the semantic similarity between the references and candidates. However, in an open-ended generation, a generated text of high quality could have a different meaning with the reference.

Recently, various learned metrics are also proposed. Blend \citep{blend_metric} uses an SVM model to combine different existing evaluation metrics. \ruse \citep{ruse_metric} evaluates machine translation by training sentence embeddings on data obtained in other tasks. \citet{cui_imagecaption_evaluation} trains a neural network conditioning on image to distinguish between human and machine-generated captions. \bleurt utilizes a BERT model fine-tuned on various automated metrics to evaluate the quality of the generated text. 
\citet{comparator_evaluator} propose to evaluate
NLG models by learning to compare a pair of generated sentences by fine-tuning \bert.

Since human evaluation remains the gold standard for almost all NLG tasks, another popular research topic is to train an automated metric based on human evaluation.
To calibrate human judgments and automatic evaluation metrics, model-based approaches that use
human judgments as attributes or labels have been proposed.
HUSE \citep{huse-metric} connects statistical evaluation with the human evaluation to evaluate summarization and chit-chat dialogue. \citet{adem} train an evaluation model based on abundant human judgment to score dialog response. These methods require heavy human annotation as supervision and risk poor generalization to new domains. Meanwhile, our method does not require human labeling and applies to generate open-ended generation tasks.
\section{Method}
\label{sec:method}

In this section, we first present our proposed metric \method and uncertainty estimation, then we introduce the overall evaluation process. 

\subsection{\method}
\label{sec:perception-score}
\begin{figure*}[h!]
    \centering
    \includegraphics[scale=0.25]{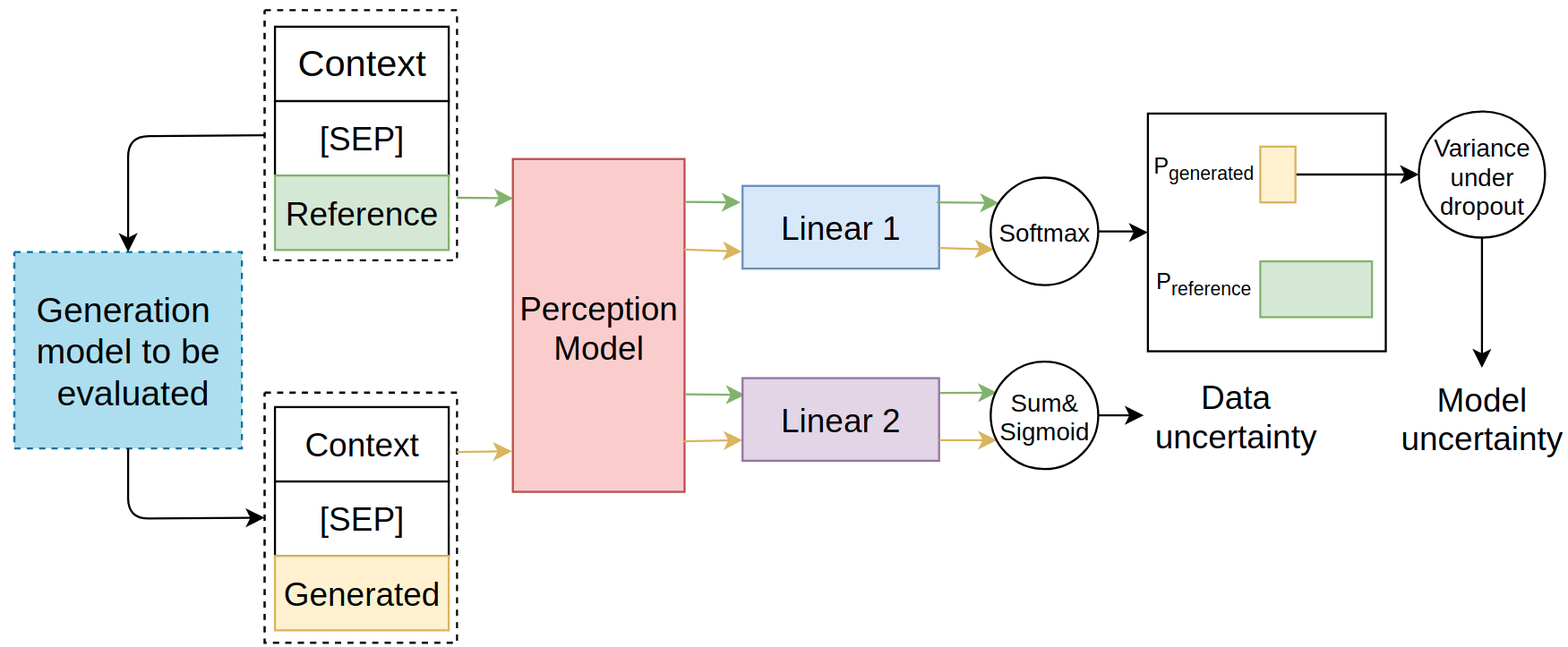}
    \caption{The main structure of \method measuring process. We first get the generation from the generation model to be evaluated. Then we enhance the representation of context and generation by incorporating context. The Perception Model is a pre-trained model with a strong understanding capacity. A softmax function is applied to bound \method between (0,1). Data uncertainty is the output of the neural network. Model uncertainty is the variance of $P_{generated}$ under difference network dropout settings.}
    \label{fig:model-structure}
\end{figure*}


For a generation system to be evaluated, denote its generation on the dataset as 
$\generation{} = \langle \senthyp{}_1, \dots, \senthyp{}_n  \rangle$, where $\senthyp{}_i$ is the generation for $i$-th sample in the dataset, and denote the corresponding reference as $\Reference{} = \langle \sentref{}_1, \dots, \sentref{}_n  \rangle$. An automated metric evaluates the quality of the generation system by scoring $\generation{}$ with $\Reference{}$ as the goal standard. 

Different metrics usually focus on a different aspect in the scoring process. For example, \bleu focuses on word surface overlapping, and \bertscore focuses on token-level semantic similarity. A qualified generation usually meets various criteria and high performance in one aspect does not guarantee quality, especially in open-ended natural language generation tasks.

We define Perception Score as a function $\delta$ that measures the realness of the generation with the reference as the goal standard. Since realness is a relative measure, in contrast to the static evaluation metrics, \method is a learning-based metric and should be parameterized to fit various evaluation tasks.

\begin{figure}
    \centering
    \includegraphics[scale=0.13]{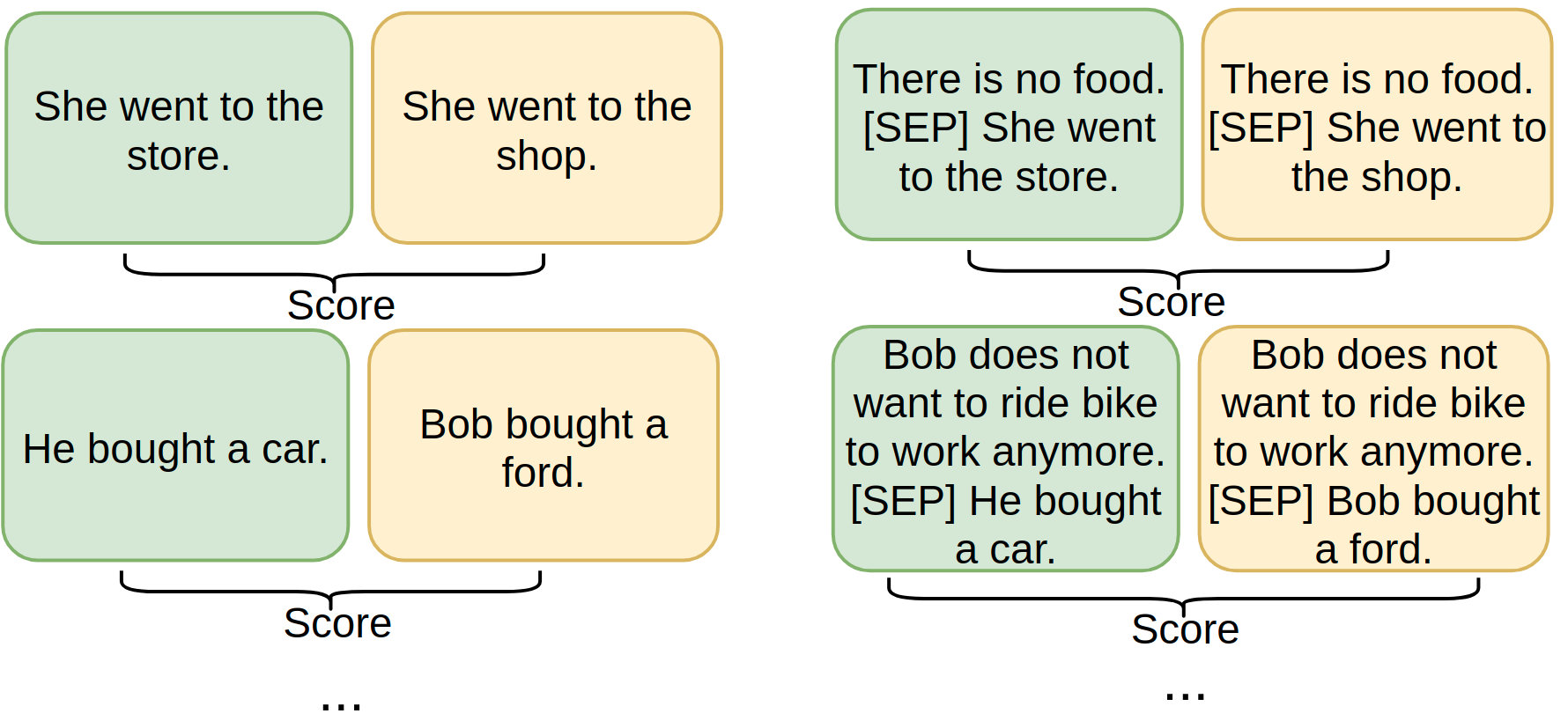}
    \caption{Previous methods (left) do not consider information context, \method (right) takes context into consideration and therefore the semantic meaning is more comprehensive.
}
    \label{fig:my_label}
\end{figure}

We define this realness by aggregating the inter-distances between every pair of samples in the reference set and generation sets.
In mathematics terms, such realness measure is defined as:

\begin{equation}
    \delta_\theta = \argmax_\theta \mathbb{E}_{\senthyp{}\sim \mathbb{P}_\generation, \sentref{}\sim \mathbb{P}_\Reference}   [\delta_\theta (\senthyp{} , \sentref{})]
    \label{eq:delta}
\end{equation}

Since in many NLG tasks, the semantic meaning of the generation is context-dependent, we further incorporate context information into $\generation{}$ and $\Reference{}$ to get comprehensive understanding in evaluation process.
We now have $\generation{}$ as $\langle \{\context{}_1, \senthyp{}_1\}, \dots, \{\context{}_n, \senthyp{}_n\}  \rangle$, and have $\Reference{}$ as $\langle \{\context{}_1, \sentref{}_1\}, \dots, \{\context{}_n, \sentref{}_n\}  \rangle$, where $c_i$ is the context of $i$-th sample in the dataset.

Eq~\ref{eq:delta} can be viewed as one form of Earth Mover's Distance (EMD). EMD is a measure of the distance between two probability distributions.
The goal is to compute a reasonable and efficient approximation of EMD.

Apply the Earth-Movers' Distance or Wassertein-1, we have 
\begin{equation}
    W(\mathbb{P}_\generation, \mathbb{P}_\Reference) = \underset{\gamma \in \prod (\mathbb{P}_\generation, \mathbb{P}_\Reference)}{\mathrm{inf}} \mathbb{E}_{(\sentref{},\senthyp{}) \sim \gamma}[ ||\sentref{} - \senthyp{}||]  
\end{equation}
where $\prod(\mathbb{P}_\generation, \mathbb{P}_\Reference)$ denotes the set of all joints distributions.
$\gamma$ indicates how much ``mass" must be transported from $\senthyp{}$ to $\sentref{}$ in order to transform the distribution $\mathbb{P}_\generation$  to the distribution $\mathbb{P}_\Reference$.
The EM distance is the optimal transport plan.

The formula
Following \citet{wgan}, we convert the formula by using the supremum over the 1-Lipschitz functions.

\begin{equation}
\label{eq:sup}
   W(\mathbb{P}_\generation, \mathbb{P}_\Reference) = \underset{|f|_L \leq 1}{\mathrm{sup}}  \mathbb{E}_{\senthyp{} \sim \mathbb{P}_\generation} [f(\senthyp{})] - \mathbb{E}_{\sentref{} \sim \mathbb{P}_\Reference} [f(\sentref{})]
\end{equation}

When we use a neural network to find the solution, it is equivalent to optimize the following problem:

\begin{equation}
    W(\mathbb{P}_\generation, \mathbb{P}_\Reference) = \underset{\theta}{\mathrm{max}} \, \mathbb{E}_{\senthyp{} \sim \mathbb{P}_\generation{}, \sentref{} \sim \mathbb{P}_\Reference} [f_\theta(\senthyp{}) - f_\theta(\sentref{})]
    \label{eq:max}
\end{equation}

with a gradient penalty loss to fulfill the Lipschitz constrain:
\begin{equation}
    GP = \underset{\senthyp{} \sim \mathbb{P}_{\generation{}}, \sentref{} \sim \mathbb{P}_{\Reference{}}}{\mathbb{E}}
    [(||\nabla_{\senthyp{}, \sentref{}}D(\senthyp{}, \sentref{})||_2-1)^2]
\end{equation}

However, one problem is that in text space, the elements are all discrete.
Sequences can also have various lengths, unlike images where the resolution is fixed at the beginning.
Then there exists a much larger space for fake sequences than the real sequences.
To resolve this problem, we bound the maximum earth-mover distance to be $1$. Otherwise, the original distance measure diverges quickly in text space.
Thus, instead of directly using the output from Perception Model as the \method, we normalize the output with softmax.
We denote the output of the Perception model of reference sample and generation sample as $f_\theta'(x)$ and $f_\theta'(y)$, then we have $f_\theta(x)$ and $f_\theta(y)$,

\begin{equation}
    P_{generated} = f_\theta(x) = \frac{e^{f_\theta'(x)}}{e^{f_\theta'(x)} + e^{f_\theta'(y)}} 
\end{equation}

and

\begin{equation}
    P_{reference} = f_\theta(y) = \frac{e^{f_\theta'(y)}}{e^{f_\theta'(x)} + e^{f_\theta'(y)}} 
\end{equation}

Then we get the final optimization object for neural network:

\begin{equation} \label{eq1}
\begin{split}
   \underset{\theta}{\mathrm{argmax}} \, \ \mathbb{E}_{x \sim \mathbb{P}_r, y \sim \mathbb{P}_g}{[\mathrm{log} \ P_{reference}]}
\end{split}
\end{equation}

\subsection{Uncertainty}

Uncertainty comes with judgment. For example, when we consider a story to be good, we also know how confident we are to the judgment. When evaluating a generation system in NLG tasks, weighing the generation that could be scored with high confidence more leads to a better evaluation of the system overall. Since \method is based on neural networks, we could use the feature of neural networks to measure the uncertainty.
Uncertainty in the neural network includes data uncertainty and model uncertainty \citep{gal2016uncertainty}. We utilize both of them for a more accurate evaluation result.

\citet{model_uncertainty} proposes using dropout to measure model uncertainty. During the evaluation process, different random dropout settings are used to get different results from the neural network. Then the variance of those results is a measure of the model uncertainty. We adapt that idea and use dropout to measure the model uncertainty of \method. When $m=1$, \method is sure that its scoring is correct.
\begin{equation}
    m = 1- \Var(P_{generated})
\end{equation}

As for data uncertainty, following \citet{data_uncertainty}, we use additional fully-connected layers and sigmoid function.

\begin{equation}
c = f_{\theta}(x, y). \quad x\in\mathbb{P}_g, y\in\mathbb{P}_r 
\end{equation}

Then the optimize object is adjusted by interpolating
between the original \method and the data uncertainty:
\begin{equation}
p'  = c \cdot P_{reference} + (1 - c).\
\end{equation}

And instead of optimizing $P_{reference}$, we optimize $p'$:
\begin{equation}
\mathcal{L}_{t} = -\sum\limits_{i=1}^M \log(p'_{i}) .\
\end{equation}

As suggested by \citet{data_uncertainty}, we utilize an extra regularizer serving as a loss to prevent the network from minimizing the loss by
always choosing data uncertainty as 0. 

\begin{equation}
\mathcal{L}_{c} = -\log(c) .\
\end{equation}

The final loss is the weighted sum of the losses we mentioned:
\begin{equation}
\mathcal{L} = \mathcal{L}_{t} + \lambda\mathcal{L}_{c} + \beta GP.\
\end{equation}

\subsection{Evaluation Process}
\label{subsec:evaluation-process}

Now we introduce how to use \method in NLG tasks. The overall using process is similar to the general neural network training process. The dataset is the concatenation of $\generation{}$ and $\Reference{}$. $\generation{}$ is generated by the generation system that we want to evaluate. A Perception Model such as \roberta{} learns to grasp the criteria of a high quality generation and to find the difference between the references and the candidates. Then the \method is the result of the test dataset. Unlike a general classification task, where a more powerful model usually leads to higher accuracy, in our evaluation process, a powerful model leads to a more close approximation of the EMD between $\generation{}$ and $\Reference{}$. In other words, the highest possible \method is only related to the difference between $\generation{}$ and \Reference{}.

During the test time, as introduced before, we weigh each generation with it's corresponding uncertainty. Then the \method for $\generation{}$ is:
\begin{equation}
    P_{sys} = \sum^{n}_{i=1} w_{i} * P^{i}_{generated}
\end{equation}
where $w$ is calculated by model uncertainty $m$ and data uncertainty $c$.
\begin{equation}
    w_i = \frac{1/(c_i + m_i)}{\sum_{k=1}^{n}1/(c_k + m_k)}
\end{equation}

Higher $P_{sys}$ means higher system generation quality.
A system whose generation is of the same quality of the reference should get $P_{sys}$ around 0.5, meaning the Perception Model can not distinguish them by text quality.

\section{Experiment Setting}
\label{sec:exp}

A qualified metric should provide similar results with human judgments.
We evaluate \method in both open-ended conditional generation and open-ended unconditional generation. The Perception Model could be any neural networks with a strong understanding ability. We use \roberta{} in this paper.
We choose the state-of-the-art generation model GPT-2 with different architectures and training hyperparameter choices as the system to be evaluated by \method.
We test our proposed metric on four open-ended natural language generation tasks, including two conditional generation tasks and two unconditional generation tasks.
Refer Appendix~\ref{sec:sup_data} for detail about the datasets. 
 
\subsection{Conditional Generation}
In an open-ended conditional generation, a system needs to generate text based on a given context.
We first evaluate \method on \rocstory Completion task \citep{rocstory_dataset}. The purpose of this task is to generate an open-ended ending for a four-sentence short story.
We then evaluate our method on \imdbCon \citep{imdb_dataset}. The purpose of this dataset is to finish the review based on a movie review context. Since both are open-ended generation tasks, the generation from the language system could share little semantic similarity with the given reference in the dataset.

\subsection{Unconditional Generation}
We also evaluate \method on two unconditional generation tasks \imdbUn{} \citep{imdb_dataset} and \coco{} \citep{coco_dataset}. The purpose of \imdbCon{} is to generate an original and high-quality movie, and the purpose of \coco{} is to generate high-quality image caption unconditionally. Unlike in conditional generation, there is no fixed reference for a system generation. Other metrics such as \bleu need to set all text in the dataset as references. However, since \method scores based on how much the generation fits the task and is not limited to word-surface or semantic level similarity, a few references would be enough for \method to measure a generation. In the experiment, we use four references for each 
generation and take the average \method.

\section{Result}
\label{sec:result}

\begin{table*}[h!]
    \centering
    \begin{tabular}{ p{4.5cm}||p{1.7cm}|p{1.7cm}|p{1.7cm}|p{1.7cm}}
     \hline
     Metrics & R & LMRC &  CIC &LMRU\\
     \hline
     \bleu-1   & 0.195    &0.386    & 0.504  &   0.119 \\
     \hline
     \bleu-2   & 0.491   & 0.316  &  0.0933 & 0.0908\\
     \hline
     \bleu-3  & 0.432  & -0.485  & 0.116   & 0.152\\
     \hline
     \bleu-4 &  0.355  & -0.390  &  0.0523  & 0.173\\
     \hline
     \perplexity & 0.638  & 0.419  &   0.4948 &   0.128  \\
     \hline
     \bertscore (base) &  0.290    &  0.454   & -    & - \\
     \hline
     \bertscore (large) & 0.282   & 0.474  &  -  & - \\
     \hline
     \moverscore (base) & 0.313 & 0.467  &  - & - \\
     \hline
     \moverscore (large) & 0.328 & 0.487  &  - & - \\
     \hline
     \bleurt (base) & 0.513    &  0.467    & -    & - \\
     \hline
     \bleurt (large) & 0.529 & 0.491 & - & - \\
     \hline
     \method (base)&  0.671   & 0.488   & 0.563   & 0.231\\
     \hline
     \method (large) &  \textbf{0.692}   & \textbf{0.494}   & \textbf{0.578}   & \textbf{0.249}\\
     \hline

    \end{tabular}
    \caption{Correlation with human judgement. R is short for \rocstory{} dataset. LMRC is short for \imdbCon{} dataset; CIC is short \coco{} dataset; LMRU is short for \imdbUn{} dataset. Perception Score outperforms other metrics by a large margin in all four tasks.}
    \label{tab:correlation}
\end{table*}
\subsection{Correlation with Human Judgement}

The performance of an evaluated metric is usually measured by its correlation with the human judgment \citep{evaluation-survey}. Following previous work \citep{bertscore,moverscore}, we use Turing score M1 as the human judgment. Turing score M1 is the percentage of a model’s generations that are evaluated as better or equal to the references. We compare with various popular metrics including: \bleu, \perplexity, \bertscore, \moverscore, \bleurt.
Following \bertscore, we compute the Pearson correlation with Turing score M1. 

As is shown in Table~\ref{tab:correlation}, \method has the highest correlation with human judgment among all metrics across all the four tasks. The popular n-gram metric \bleu show a low correlation with human judgments. We noticed that in the conditional generation tasks, the word level overlap constitutes a considerable ratio of a meaningless word such as pronoun and Be Verb. \bleu score suffers from low overlap in the open-ended conditional generation.  

\method also outperforms other transformer-based metrics \bertscore, \moverscore, and \bleurt by a large margin. Since there is no corresponding reference for each generation in unconditional generation tasks, researchers usually utilize all samples in the dataset as references. 
However, during the experiment, we found that the same strategy does not apply to \bertscore, \moverscore, and \bleurt in unconditional generation task. The calculation of those metrics involves the forward calculation process of a large neural model such as \roberta{} or \bert{}, thus it is almost impossible to do that calculation with all samples in the dataset. Meanwhile, \method can be used to evaluate unconditional generation tasks since it only requires one sample as a reference because it measures the quality of the generation instead of measuring the similarity between the generation and the reference. 
For these transformer-based metrics, we compare both base size and large size for completeness. Note that \method of a base version outperforms \bertscore and \moverscore of a large version.

\subsection{Ablation Study}
\begin{table}[h!]
    \centering
    \begin{tabular}{ p{4.4cm}||p{1.0cm}|p{1.0cm}}
     \hline
       & R & CIC \\
     \hline
     \method (base) &  0.671   & 0.563 \\
     w/o m & 0.653   & 0.551\\
     w/o c & 0.636   & 0.538\\
     \hline

    \end{tabular}
    \caption{Ablation study on uncertainty. R is short for \rocstory{} LMRC is short for \imdbCon{}; CIC is short \coco{}; }
    \label{tab:ablation-study}
\end{table}
We conduct an ablation study on \rocstory{} and \coco{} to study the influence of uncertainty. Table~\ref{tab:ablation-study} presents the experiments results. We found that data uncertainty contributes more to the evaluation performance than the model uncertainty. Note \method still outperforms other metrics without using uncertainty to re-weigh each generation.

\subsection{Robustness Analysis}

We also tested the robustness of \method. After a Perception Model is trained, it also learns the standard of good generation. For example, when it is trained to find the realness of real stories, it also learns what qualifies a good story instead of overfitting the dataset.
We first created four kinds of story endings on the test set of \rocstory{} dataset, and then evaluate these endings with various generation metrics. The created endings types are as following: 1) Human written ending (HWE). Given the story context, turkers created a reasonable story ending. This is used to test metric performance for high quality generation. 2) Lexical different human written ending (LDHWE). Given the story context and the story ending, turkers created a reasonable story ending while avoiding using words from the original reference.  3) Human writing ending with better quality (HWEBQ). Given the story context and the story ending reference, turkers created a story ending that is of better quality than the original reference. To reduce the bias, another group of turkers will make sure the created endings is of better quality than the original ones. 4) Adversarial endings (AE). Given the context and the reference, turkers modify the original reference as little as possible to create an unreasonable story ending. 
we randomly selected 100 stories from the test set of \rocstory and ask workers on Amazon Mechanical Turk (also known as Turkers) to create these four kinds of endings.
\begin{figure}
    \centering
    \includegraphics[scale=0.13]{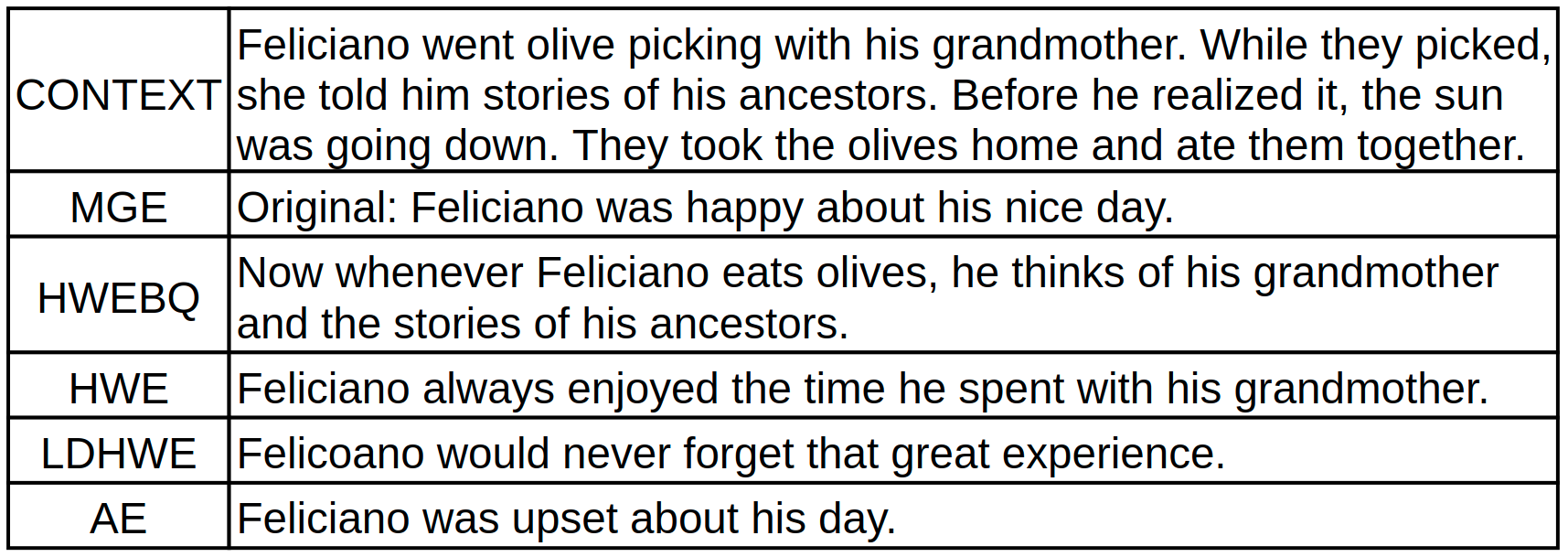}
    \caption{Created ending example used in robustness analysis.}
    \label{fig:my_label}
\end{figure}

\begin{table*}[h!]
    \centering
    \begin{tabular}{ p{3.5cm}||p{1.5cm}|p{1.5cm}|p{1.5cm}|p{1.5cm}|p{1.5cm}}
     \hline
     Metric & MGE & HWE & LDHWE & HWEBQ & AE \\
     \hline
     \bleu-1 & 0.1551 & 0.1348 & 0.1406 & 0.1612  & 0.6816\\
     \hline
     \bleu-2 & 0.0468 & 0.0359 & 0.0309 & 0.0514 & 0.5809\\
     \hline
     \bleu-3 & 0.0142 & 0.0125 & 0.0104 & 0.0168 &0.4411\\
     \hline
     \bleu-4 & 0.0052 & 0.0054 & 0.0038 & 0.0074 &0.3014\\
     \hline
     \bertscore & 0.4223 & 0.3631 & 0.3742  & 0.3974 & 0.7586 \\
     \moverscore & 0.2439 & 0.2220 & 0.2359 & 0.2510 & 0.5935 \\
     \hline
     \bleurt & 0.4785 & 0.4847 & 0.4768 & 0.5007 & 0.6458\\
     \hline
     \method & 0.1115 & 0.4817 & 0.6001 & 0.7845  & 0.3650\\
     \hline
    \end{tabular}
    \caption{Robustness analysis results on \rocstory dataset. MGE is short for model generated ending; HWE is short for Human written ending; LDHWE is short for lexical different human written ending; HWEBQ is short for human writing ending with better quality; AE is short for Adversarial endings.}
    \label{tab:adversarial-attack}
\end{table*}
The evaluation result is shown in Table~\ref{tab:adversarial-attack}. We also show the various metric scores on the generated ending (GE). 
Since \bertscore has a small scale between 0.8~1.0, \moverscore could give a negative score, and \bleurt scores ranges from negative number to more than 1.0, so we re-scale them to 0.0~1.0. The scaling does not affect the ranking and the Relationship between data, but only increases readability. The second column is the performance on the original dataset and shows the quality of generated story ending.

As is shown in the third column, we found that all the baseline metrics has a difficulty in recognizing the quality of the good story endings. Since all other metrics are measuring similarity in in word or semantic level, it will fail to judge a good generation that has little similarity with the given reference fairly, which is a common scenario in open-ended generation tasks. In the LDHWE column, we get similar results with the HWE column, It shows that 
whether the turkers are intended to write a ending that is different than the original reference, the overall wording and semantic of created ending is different than the original one. It matches the common sense that there could be many ground truth text in open-ended task, and using similarity based metric on limited reference will inevitably leads to quality misjudgment. In the HWEBQ column, each metrics show the superiority to the generated ending in quality by presenting a higher score than its corresponding one for the generated endings. However, there is a huge gap between the real quality difference and the quality difference shown by these baseline metrics. For example, \bleu-1 only increases 0.0061. Meanwhile, our \method successfully reflects the quality difference. It does not suggest that \method could be used to judge whether a generation is of higher quality than the given reference, but a reflection of the generalizability of \method. In the last column, we can see that all other base shows a significant boost and gives a high score for a unreasonable ending. Since \method scores much less than 0.5, showing \method scores not solely on the lexical or semantic feature, but also consider the appropriateness under the related context. It shows the necessarity of taking context into consideration in open-ended generation tasks.

\section{Conclusion}
\label{sec:conclusion}

We propose a learned metric \method for text generation tasks. \method evaluates the semantic difference between generations and references and consider the context information. \method show superiority than various metrics with extensive experiments. 

\clearpage

{\footnotesize
\bibliographystyle{aaai}
\bibliography{bibliography}
}

\clearpage
\appendix
\onecolumn
\section{Dataset}
\label{sec:sup_data}

Four different datasets are used. For conditional generation task, we use \rocstory dataset \citep{rocstory_dataset} and IMDB review conditional dataset. \rocstory is to generate an open-ended ending for a four-sentence story.
Unlike unconditional or other more constrained text generation tasks, remembering the training set will not lead to good performance in this task, as every story is unique.
IMDB review conditional dataset came from Large Movie Review Dataset v1.0. For each review in the Large Movie Review Dataset, we set the context before the last sentence as context and set the last sentence as a review ending.
For unconditional generation task, we use \coco dataset \citep{coco_dataset} and IMDB review unconditional dataset. IMDB review came from Large Movie Review Dataset v1.0. 

\begin{table*}[!h]
    \centering
    \begin{tabular}{ p{5.0cm}||p{1.3cm}|p{1.9cm}|p{1.9cm}}
     \hline
     Dataset Name & Train Tokens& Dev Tokens & Test Tokens \\
     \hline
     \rocstory &    98,162& 1,871& 1,871   \\
     \hline
     IMDB conditional & 21,730 &10,854 &10,854  \\
     \hline
     COCO Image Caption & 10,000& - & 10,000   \\
     \hline
     IMDB unconditional & 22,146& 11,063 & 11,062   \\
     \hline
    \end{tabular}
    \caption{Dataset statistics.}
    \label{tab:dataset_stat}
\end{table*}

\end{document}